\documentclass[runningheads,a4paper]{llncs}
\usepackage{import}

\usepackage{ifthen}
\usepackage{url}
\usepackage{amssymb}
\usepackage{amsmath}
\usepackage{import}
\usepackage{xparse}
\usepackage{amsmath}
\usepackage[usenames,dvipsnames]{xcolor}
\usepackage{xspace}
\usepackage{datatool}
\usepackage{glossaries}

\newcommand{\m}[1]{\ensuremath{#1}\xspace}

	\newcommand{\lrule}{\m{\leftarrow}}
	\newcommand{\cause}{\m{\stackrel{c}{\lrule}}}

	\newcommand{\struct}{\m{I}}

	\NewDocumentCommand\inter{g+g}{%
	  \IfNoValueTF{#1}
	    {\struct}
	    {\m{#1^{#2}}}}

	\newcommand{\bool}{\m{\mathbb{B}}}

	\renewcommand{\int}{\m{\mathbb{Z}}}

	\NewDocumentCommand\subs{g+g}{%
	  \IfNoValueTF{#1}
	    {\m{/}}
	    {\m{#1/ #2}}}

	\newcommand{\logicname}[1]{\text{\sc #1}\xspace}

	\newcommand{\idp}{\logicname{IDP}}

	\newcommand{\minisatid}{\logicname{MiniSAT(ID)}}

	\newcommand{\fodot}{\logicname{FO(\ensuremath{\cdot})}}

\newcommand{\ouracronym}[3]{%
	\newacronym{#1}{#2}{#3}
	\expandafter\newcommand\csname #1\endcsname{\gls{#1}\xspace}%
}
	\ouracronym{FO}{FO}{first-order logic}
	\ouracronym{MX}{MX}{Model Expansion}
	\ouracronym{MO}{MO}{Model Optimization}
	\ouracronym{ASP}{ASP}{Answer Set Programming}
	\ouracronym{TP}{TP}{Theorem Proving}
	\ouracronym{LP}{LP}{Logic Programming}
	\ouracronym{CP}{CP}{Constraint Programming}
	\ouracronym{FP}{FP}{Functional Programming}
	\ouracronym{KR}{KR}{Knowledge Representation}
	\ouracronym{CSP}{CSP}{Constraint Satisfaction Problem}
	\ouracronym{SMT}{SMT}{SAT Modulo Theories}
	\ouracronym{KBS}{KBS}{Knowledge Base System}
	\ouracronym{NNF}{NNF}{Negation Normal Form}
	\ouracronym{FNNF}{FNNF}{Flat Negation Normal Form}
	\ouracronym{DefNNF}{DefNNF}{Definition Negation Normal Form}
	\ouracronym{CDCL}{CDCL}{Conflict-Driven Clause-Learning}
	\ouracronym{LCG}{LCG}{Lazy Clause Generation}

	\makeatletter
	\def\ifenv#1{
	\def\@tempa{#1}%
	\def\@ttempa{#1*}%
	\ifx\@tempa\@currenvir
	\expandafter\@firstoftwo
	\else
	\expandafter\@secondoftwo
	\fi
	}
	\makeatother

	\newcommand{\ddrule}[4]{\ensuremath{#1 \leftarrow #2 & \{#3\} & #4}}
	\newcommand{\drule}[2]{\ensuremath{#1 & \leftarrow & #2}}

	\newcommand{\darule}[4]{\ensuremath{#1 \leftarrow #2 & \{#3\} & #4}}
	\newcommand{\arule}[2]{\ensuremath{#1 \, &\leftarrow \, #2}}

	\newcommand{\LNDRule}[2]{
	\ifenv{array}
	{\drule{#1}{#2}}
	{ \ifenv{align}
		{\arule{#1}{#2}}
		{\ifenv{align*}
		{\arule{#1}{#2}}
		{ERROR: using LDRule in unsupported environment: \@currenvir}
		}
	}
	}

	\newcommand{\LDRule}[4]{
	\ifenv{array}
	{\ddrule{#1}{#2}{#3}{#4}}
	{ \ifenv{align}
		{\darule{#1}{#2}{#3}{#4}}
		{\ifenv{align*}
		{\darule{#1}{#2}{#3}{#4}}
		{ERROR: using LDRule in unsupported environment: \@currenvir}
		}
	}
	}

	\NewDocumentCommand\LRule{m+g+g+g}{%
		\IfNoValueTF{#2}%
		{#1.&}{%
		\IfNoValueTF{#3}
		{\LNDRule{#1}{#2.}}
		{\LDRule{#1}{#2.}{#3}{#4}}%
		}
	}

	\NewDocumentCommand\CLRule{m+g}{%
	\ifenv{array}
	{\cdrule{#1}{#2}}
	{ \ifenv{align}
		{\carule{#1}{#2}}
		{\ifenv{align*}
			{\carule{#1}{#2}}
			{ERROR: using CLRule in unsupported environment: \@currenvir}
		}
	}
	}

	\NewDocumentCommand\carule{m+g}{%
		\IfNoValueTF{#2}
			{\ensuremath{#1.}}
			{\ensuremath{#1 \, &\cause \, #2}}}
	\NewDocumentCommand\cdrule{m+g}{%
		\IfNoValueTF{#2}
			{\ensuremath{#1.}}
			{\ensuremath{#1 & \cause & #2}}}

	\newcommand{\algrule}[4]{
	\hbox{{#1}:}& 
	\quad #2 ~\longrightarrow~ #3 
	\hbox{~ if } #4\\
	}

	\newcommand{\AlgoRule}[4]{
	\ifenv{array}
	{\algrule{#1}{#2}{#3}{#4}}
		{ERROR: using AlgoRule in unsupported environment: \@currenvir}
	}

\newcommand{\commentstyle}{\color{Gray}}
	\usepackage[final]{listings}

	\lstdefinelanguage{idp}{
		morekeywords=[1]{namespace,vocabulary,theory,structure,procedure,term,set,formula, spec, specification},
		morekeywords=[2]{include,using,type,isa,contains,partial,extern,LFD,GFD,constructed,from,constraint,func,pred,supertype,of,subtype,define},
		morekeywords=[3]{int,float,char,string,nat},
		morekeywords=[4]{if,then,else,for,end},
		morecomment=[s]{/*}{*/},	
		morecomment=[l]{//}
	}
	\newcommand{\ignore}[1]{}

	\newboolean{nocomments}
	\setboolean{nocomments}{false}

	\newboolean{commentmargin}
	\setboolean{commentmargin}{true}

	\newcommand{\namedcomment}[3]{
		\ifthenelse{\boolean{nocomments}}
		{} %IF no comments, write nothing
		{ %Otherwise
			\ifthenelse{\boolean{commentmargin}}
				{ {\color{#3} \marginpar{\color{#3}\sc #2}#1}  } %Name in margin
				{  {\color{#3} {\sc #2}: #1}  } %Name not in margin
		}
	}
	\newcommand{\mnamedcomment}[3]{\ifthenelse{\boolean{nocomments}}{}{{\marginpar{ \color{#3}{\sc #2}:#1}}}}

	\newcommand{\jo}[1]{ \namedcomment{#1}{jo}{Fuchsia}}

	\usepackage{soul}

\usepackage{etoolbox}

\newcommand\setcitation[2]{%
  \csdef{mycommoncitation#1}{#2}}

\setcitation{IDP}{WarrenBook/DeCatBBD14}
\setcitation{fodot}{tocl/DeneckerT08}
\setcitation{CP}{Apt03}
\setcitation{EZCSP}{lpnmr/Balduccini11}
\setcitation{KR}{Baral:2003}
\setcitation{ASPComp2}{lpnmr/DeneckerVBGT09}
\setcitation{ASPComp3}{journals/tplp/CalimeriIR14}
\setcitation{ASPComp4}{conf/lpnmr/AlvianoCCDDIKKOPPRRSSSWX13}
\setcitation{CPSupport}{ictai/DeCat13}
\setcitation{functionDetection}{iclp/DeCatB13}
\setcitation{fodot2asp}{DeneckerLTV12} %TODO replace by journal publication if one is publishedr
\setcitation{Tarskian}{DeneckerLTV12} %Same as the one above 
\setcitation{Inca}{iclp/DrescherW12}
\setcitation{csp2asp}{ijcai/DrescherW11}
\setcitation{DPLLT}{cav/GanzingerHNOT04}
\setcitation{AspInPractice}{synthesis/2012Gebser}
\setcitation{clasp}{ai/GebserKS12}
\setcitation{oclingo}{kr/GebserGKOSS12}
\setcitation{gringo}{lpnmr/GebserST07}
\setcitation{cmodels}{aaai/GiunchigliaLM04}
\setcitation{inputster}{tplp/Jansen13}
\setcitation{DLV}{tocl/LeonePFEGPS06}
\setcitation{LearningPaper}{TPLP/BruynoogheBBDDJLRDV} %TODO replace by published version
\setcitation{clog}{iclp/BogaertsVDV14} %TODO replace by journal publication if one is published
\setcitation{foc}{iclp/BogaertsVDV14}
\setcitation{inferenceClog}{ecai/BogaertsVDV14}
\setcitation{examplesClog}{nmr/BogaertsVDV14b} %TODO replace by better publication if possible
\setcitation{AFT}{DeneckerBV12}
\setcitation{KBS}{iclp/DeneckerV08}
\setcitation{KBPE}{inap/DePooterWD11}
\setcitation{lazyGrounding}{corr/CatDSB14} %TODO replace by published version
\setcitation{ASP}{marek99stable}
\setcitation{satid}{sat/MarienWDB08}
\setcitation{lazyclausegeneration}{constraints/OhrimenkoSC09}
\setcitation{FP}{ACMCS/Hudak89}
\setcitation{GroundingWithBounds}{jair/WittocxMD10}
\setcitation{SAT}{faia/SilvaLM09}
\setcitation{LTC}{iclp/Bogaerts14}
\setcitation{SPSAT}{ictai/DevriendtBMDD12}
\setcitation{BreakID}{cspsat/DevriendtBB14}
\setcitation{LCG}{stuckeyLCG}
\setcitation{MiniZinc}{conf/cp/NethercoteSBBDT07}
\setcitation{amadini}{cpaior/AmadiniGM13}
\setcitation{bootstrapping}{lash/BogaertsJDJBD14}
\setcitation{GroundedFixpoints}{BogaertsVD15}
\setcitation{LogicBlox}{datalog/GreenAK12}
\setcitation{proB}{journals/sttt/LeuschelB08}
\setcitation{NaturalInductions}{KR/DeneckerV14} %TODO replace by journal publication if one is published
\setcitation{LP}{jacm/EmdenK76}
\setcitation{SMT}{faia/BarrettSST09}
\setcitation{AF}{ai/Dung95}
\setcitation{ADF}{kr/BrewkaW10}
\setcitation{}{}
\setcitation{}{}
\setcitation{}{}
\setcitation{}{}
\setcitation{}{}
\setcitation{}{}
\setcitation{}{}
\setcitation{}{}
\setcitation{}{}
\setcitation{}{}
\setcitation{}{}
\setcitation{}{}
\setcitation{}{}
\setcitation{}{}
\setcitation{}{}
\setcitation{}{}
\setcitation{}{}
\setcitation{}{}
\setcitation{}{}
\setcitation{}{}
\setcitation{}{}
\setcitation{}{}
\setcitation{}{}
\setcitation{}{}
\setcitation{}{}
\setcitation{}{}
\setcitation{}{}
\setcitation{}{}
\setcitation{}{}
\setcitation{}{}
\setcitation{}{}
\setcitation{}{}
\setcitation{}{}
\setcitation{}{}
\setcitation{}{}
\setcitation{}{}
\setcitation{}{}
\setcitation{}{}
\setcitation{}{}
\setcitation{}{}
\setcitation{}{}
\setcitation{}{}
\setcitation{}{}
\setcitation{}{}
\setcitation{}{}
  
\usepackage{amssymb}
\usepackage{mathrsfs}
\usepackage{color}
\usepackage{graphicx}
\usepackage{algorithm}
\usepackage{subfig}

\usepackage{url}
\usepackage{comment}

\urldef{\mailsc}\path|erika.siebert-cole, peter.strasser, lncs}@springer.com|

\begin{document}
\mainmatter  % start of an individual contribution

%\title{A MIP Backend for the Optimization Problems in \fodot, a Logic Based Specification Language}
\title{A MIP Backend for the \idp System}
\author{San Pham \and Jo Devriendt \and Maurice Bruynooghe \and Patrick De Causmaecker}
\institute{Department of Computer Science, KU Leuven}
\titlerunning{A MIP Backend for \idp}
\maketitle

\begin{abstract}
%The \idp system is a \emph{knowledge base system} that currently solves its combinatorial optimization problems with the backend constraint programming (CP) solver \minisatid. However, mixed integer programming (MIP) solvers are designed to solve the same combinatorial optimization problems, so a MIP backend to \idp could outperform the current CP solver on certain problems. This paper investigates how the MIP solver CPLEX can be used as a backend for \idp, and it experimentally compares the performance of both the CP and MIP approach. These experiments indicate that a MIP backend would greatly improve IDP's performance on a wide array of problems.

The \idp \emph{knowledge base system} currently uses
  \minisatid as its backend Constraint Programming (CP) solver. A few
  similar systems have used a Mixed Integer Programming (MIP) solver
  as backend. However, so far little is known about when the MIP
  solver is preferable. This paper explores this question. It
  describes the use of CPLEX as a backend for \idp and reports on
  experiments comparing both backends.
\end{abstract}

\section{Introduction}\label{sec:intro}
Mixed Integer Programming (MIP) provides powerful solvers for tackling
combinatorial optimization problems. Hence it looks attractive to
explore the use of such a backend for CP and ASP systems. In
\cite{minizinc2mip}, the use of a MIP backend for MiniZinc \cite{Zinc}
was explored; however, not much guidance is provided for chosing
between the MIP and the CP backend. In \cite{kr/LiuJN12}, a MIP solver
is used as a backend for an ASP
\cite{cacm/BrewkaET11,synthesis/2012Gebser} system. They conclude that
MIP offers most gain in optimization problems, and gives unsatisfying
results on search problems with only boolean variables and constraints. 

In this paper we describe the use of a MIP backend for the \idp
system
\cite{iclp/DeneckerV08,inap/DePooterWD11,TPLP/BruynoogheBBDDJLRDV}. 
As the above systems, the \idp system is a declarative system aiming
to tackle problems using an accessible high-level modeling language.

Our main contribution is to
analyze in more detail for which problems the MIP backend is to be
preferred over the native backend.  
%\san{I think our main contribution is to extend the variety of problems that IDP can handle effectively by introducing the MIP backend.}
We also pay attention to the
optimality of the translation and compare the performance of the
translation with that of hand coded MIP specifications. 
Our experiments confirm that for many optimization problems, a MIP 
backend outperforms the CP backend. However, on problems with 
certain reified sum constraints and deep planning problems a CP approach performs
significantly better, regardless of whether the problem is in fact an optimization 
problem or not.

In Section~\ref{sec:idp_intro}, we introduce the \idp system, in
Section~\ref{sec:transformation} we sketch the translation to MIP input,
in Section~\ref{sec:experiments} we report on experiments and we
conclude in Section~\ref{sec:conclusion}.
 
\section{\idp and \fodot}
\label{sec:idp_intro}
The \idp system \cite{iclp/DeneckerV08,inap/DePooterWD11} is a
knowledge base system that aims at separating knowledge from problem
solving allowing for the reuse of the same knowledge for solving different
problems.
Its input language is based on \fodot, a logic that extends
first order logic with types, aggregates, partial functions and
inductive definitions. For the latter, the meaning is based on the  
intuitive mathematical reading of definitions, and is formalized by the
\emph{parametrized well-founded semantics}
\cite{tocl/DeneckerT08,KR/DeneckerV14}. Here, we introduce
the \idp language by the example of the traveling
salesman problem domain (TSP).

\begin{example}[Traveling Salesman]
  An \idp
  specification (or model) consists of several components.
%  \jo{synchronize web-specification with paper}
  
  A first component is the \emph{vocabulary} $V$ that introduces the types,
  constants, functions and predicates used in the specification. For
  TSP, a single type $City$ is sufficient. A
  constant $Depot: City$ specifies the city that is the start and
  ending point of the trip. The network is described by a function
  $Distance(City,City):N$ with $N$ a set of natural numbers.
  The solution tour is described by a
  predicate $Next(City,City)$. Finally a predicate $Reachable(City)$
  will be used to express that all cities need to be visited starting from
  the depot.

  A second component is the \emph{structure} $S$ that describes the type and the
  input values for the instance of interest. E.g., $City=\{a,b,c\}$, $Depot=
  a$, $Distance=\{(a,a,0), (a,b,1), (a,c,2), \ldots\}$.

  A third component is the \emph{theory} $T$ that describes a solution:

  \begin{tabular}{l}
    $\forall x\colon \exists! y\colon Next(x,y).$ // for each city $x$ there
    is exactly one city $y$ next \\
    $\forall y\colon \exists! x\colon Next(x,y).$ // for each city $y$ there
    is exactly one city $x$ previous\\
    $\forall x\colon Reachable(x).$  // each city is reachable\\
    $\{\forall x\colon Reachable(x) \leftarrow x=Depot.$  // base case: the depot is reachable\\
    $\forall x\colon Reachable(x) \leftarrow \exists y\colon
    Reachable(y) \land Next(y,x). \}$ // induction step
  \end{tabular}
  
  The first two formulas express that each city is the start and end
  point of exactly one trip in the tour. Note the \emph{exists exactly} quantor
  $\exists!$ -- a shorthand for an aggregate. The third formula
  states that all cities must be reachable. The last two lines,
  between $\{$ and $\}$, inductively define the $Reachable$
  predicate. The last line states that $x$ is reachable if some $y$ is
  and $Next(y,x)$ holds.

  The fourth component specifies the \emph{term} $O$: $sum\{i~j:
  Next(i,j): Distance(i,j)\}$. $O$ represents the sum of all 
  $Distance(i,j)$ values for which $Next(i,j)$ holds, and will be used
  as objective function.

  The fifth and final component is a piece of imperative code of which
  the most important line is $result=minimize(T,S,O)[1])$
  instructing to execute the minimisation inference method with theory $T$,
  input structure $S$, optimization term $O$, and to search for the first
  optimal solution.
  
  A complete TSP specification in \idp is available at
 \url{dtai.cs.kuleuven.be/krr/idp-ide/?present=TSP}. It contains a 
 small instance that can be executed by the click of a button.
\end{example}

To solve combinatorial optimization problems, \idp currently follows a
two-phase \emph{ground-and-solve} approach.  In the first phase, an
\idp input specification is reduced to a set of constraints in
\emph{Extended Conjunctive Normal Form} (ECNF) \cite{DeCatPhd14}. This process is
comparable to the conversion of MiniZinc specifications
to Flatzinc.  The second phase consists of the actual search for an
optimal solution by calling the CP solver \minisatid.  \minisatid
iteratively searches a feasible solution for the ECNF theory,
tightening the upper bound on the objective function after each
iteration.  This loop ends when \minisatid proves that no better
solution exists, making the last feasible solution the optimal
solution. A more elaborate description and some examples can be found
in \cite{TPLP/BruynoogheBBDDJLRDV}.

\section{Transformation of \fodot to MIP}
\label{sec:transformation}
The goal of the transformation is to convert an \idp optimization specification
into a set of linear (in)equalities and the optimization term in a linear
sum. Our current transformation does not support the full \idp
language but enough to transform all specifications for which the
MIP backend looks worthwhile (so far). 

We base our transformation method on the transformation of FO to MIP
given by Williams~\cite{William09}. First, they unnest nested
formulas and instantiate FO quantors with domain elements. Second,
the resulting propositional theory is transformed using the \emph{Big
  M method}. In essence, this method transforms a constraint of the
form: 
\vspace{-1 mm}
\belowdisplayskip=1mm
\begin{align}
\label{bigM} l \Rightarrow \sum_{i}a_ix_i \geq b \quad \text{into} \quad M-Ml+\sum_{i}a_ix_i\geq b
\end{align}
The numeral $M$ is chosen big enough to make the formula
trivially satisfied when literal $l$ is false (0) while it reduces to
the linear constraint when $l$ is true (1). The exact value of $M$ is 
derived from bounds on $a_i$, $x_i$ and $b$, keeping $M$ as small 
as possible to improve linear relaxation properties of the 
transformed formula.

ECNF formulas contain two types of
variables: boolean atoms $v$ and integer constants $c$. For each of the
former, a boolean MIP variable $v^*$ is introduced and, for each of the
latter, an integer MIP variable $c^*$. Atoms $v$ usually occur as part 
of a literal $l$. We use $l^*$ to denote either the translation $v^*$ if $v$ 
occurs positively in $l$, or $(1-v^*)$ if $v$ occurs negatively in $l$.

As our transformation starts from the ECNF form, it only needs to
handle a limited set of constraints, four of them are currently supported 
by our transformation. The first type of constraints are 
\textbf{clauses} -- disjunctions of literals. 
A clause $l_1 \vee \ldots \vee l_n$ is tranformed into 
$\sum_{i} l_i^* \geq 1$.

The second type are \textbf{equivalences}, either of the form 
$v \Leftrightarrow l_1 \land \ldots \land l_n$ or of the form 
$v \Leftrightarrow l_1 \lor \ldots \lor l_n$. They are translated as 
$v \Leftrightarrow \sum_{i}l_i^* \geq n$ and
$v \Leftrightarrow \sum_{i}l_i^* \geq 1$ respectively. These forms are
further transformed with the Big M method as described above.

The third type are \textbf{reified linear sums}:
$v \Leftrightarrow \sum_{i}a_ix_i \sim b$, with $\sim$ one of $\{<,\leq,=,\geq,>,\neq\}$. 
These sums originate from linear arithmetic expressions over integer 
domains in \fodot as well as aggregate expressions over boolean 
variables. 

The strict inequalities $<$ and $>$ can be converted to $\leq$ or
$\geq$ by adding $1$ to the appropriate side of the comparison
operator, after which the formula can be transformed to a set of
formula's of form (\ref{bigM}).
The $=$ case requires the introduction of two auxiliary boolean variables $w_1,w_2$ splitting the equality into the conjunction of two inequalities
%~\cite{William09}
:
%\begin{align*}
$v \Leftrightarrow w_1 \land w_2 $, %\\
$w_1 \Leftrightarrow \sum_{i}a_ix_i \geq b $, and  % \\
$w_2 \Leftrightarrow \sum_{i}a_ix_i \leq b$.
%\end{align*}
These allow conversion to MIP via previously explained methods. 
Finally, the $\neq$ inequality is transformed to $=$ by negating both
sides of the equivalence.

Note however that the introduction of auxiliary booleans introduces a
lot of continuous solutions, reducing the so-called \emph{tightness} of the transformed constraints. For instance, in the $=$ case,
there are non-integral solutions to the translated MIP constraints such that $w_1 <1$, $w_2 <1$, $v=0$ and $\sum_{i}a_ix_i = b$. However, this violates
the original constraint that $v$ is true iff $\sum_{i}a_ix_i = b$. We
will see in the experiments that the performance of the MIP solver
suffers when many such equality constraints are present. 

%Note however that the introduction of auxiliary booleans introduces a lot of extra continuous solutions. For instance, in the $=$ case, there exist continuous values for $w_1$ and $w_2$ such that both $v=0$ and $\sum_{i}a_ix_i = b$ satisfy the MIP translation, violating the original constraint. We will later ascertain that problems with many of these equality constraints are hard for the MIP solver CPLEX. \jo{refer back to this in experiment section}

The fourth type consists of logical \textbf{definitions}.
A definition in ECNF is formed by a set of logical rules of the form:
\vspace{-3 mm}
\belowdisplayskip=1mm
\begin{align*}
%\label{prop_def}
v \leftarrow l_1 \land \ldots \land l_n \quad \text{ or } \quad
v \leftarrow l_1 \lor \ldots \lor l_n
\end{align*}
where $v$ is a boolean variable called the \emph{head}, while $l_i$ are referred to as \emph{body} literals. ECNF definitions are in so-called \emph{Definitional Normal Form}, meaning each variable occurs at most once in the head of a rule.

Our transformation follows Liu et al. \cite{kr/LiuJN12}, which uses
\emph{Clark completion} constraints \cite{adbt/Clark78} and
\emph{level mapping} constraints \cite{lpnmr/JanhunenNS09} to
translate logical rules to a simpler form.  The Clark completion
constraints simply state that head and body are equivalent (and these
equivalences are then further translated with the techniques described
above) and are sufficient for non inductive definitions. However,
inductive ones (such as the $Reachable$ definition in TSP) require also level mapping
constraints. These require for each head variable an integer variable 
representing its level. The level mapping constraints then express that a head variable can
only be made true by body literals of a lower level. As a result, the
derivation of truth is well-founded, i.e., \emph{positive loops} are
eliminated. In the TSP example in Section \ref{sec:idp_intro}, this ensures that
every city that is $Reachable$ is indeed connected to the $Depot$ and
not part of a subtour disconnected from the Depot. While  Liu et al.
require level mapping variables to be integral~\cite{kr/LiuJN12}, they can in fact be
continuous; the correctness only relies on a strict ordering. We
observed that dropping the integrality constraints on these variables
improves performance on, e.g., TSP problems.

\begin{comment}
So, for each variable $v$, a level mapping variable $z_v \geq 0$ is introduced.
A conjuctive rule $v \leftarrow l_1 \wedge \ldots \wedge l_n$ then requires the level mapping constraints
\begin{align*}
v &\Rightarrow z_v \geq z_{l_1}+ \epsilon \quad \ldots \quad v \Rightarrow z_v \geq z_{l_n}+ \epsilon
\end{align*}
where $z_{l_n}=z_{v}$ iff $v$ is the variable of $l_n$. The intuition is that if the defined variable $v$ is true, it must have a larger level mapping than all of its body variables.

For the literals $l_i$ occurring in the body of a disjunctive
rule $v \leftarrow l_1 \vee \ldots \vee l_n$, another boolean variable $w_i$ is needed. The level mapping constraints then are
\begin{align*}
v \Rightarrow &\sum_{1 \leq i \leq n} w_i \geq 1\\
w_1 \Rightarrow l_1 \quad &\ldots \quad w_n \Rightarrow l_n \\
w_1 \Rightarrow z_v \geq z_{l_1} + \epsilon \quad &\ldots \quad w_n \Rightarrow z_v \geq z_{l_n} + \epsilon
\end{align*}
stating that if $v$ is true, some $w_i$ must be true, and if some $w_i$ is true, the corresponding $l_i$ must be true while simultaneously the level mapping of $v$ should be larger than that of $l_i$.
Each of these level mapping constraints is in the form of
\ref{simple_weighted_sum}), and hence can be translated to an atomic 
constraint in the form of (\ref{translation_right}).
\end{comment}

Even though the four mentioned types of constraints suffice to experimentally analyze many interesting problems, two ECNF constraints currently are not translated to MIP. The first is the \textbf{reified conditional linear sum}, for which the terms contributing to the linear sum depend on the truth value of some literal $l_i$:
\vspace{-6 mm}
\belowdisplayskip=1mm
\begin{align}
\label{conditional_constraint}
v \Leftrightarrow \sum_{i|l_i=1}a_ix_i \sim b
\end{align}
This kind of constraint originates from aggregate expressions over sets that are not derivable from the input data. We give an unimplemented conversion of (\ref{conditional_constraint}) to MIP-transformable constraints by introducing auxiliary integer variables $x_i'$:
\vspace{-2 mm}
\belowdisplayskip=1mm
\begin{align*}
v \Leftrightarrow \sum_{i} a_ix_i' \sim b \quad | \quad \forall i\colon l_i \Rightarrow x_i' = x_i \quad | \quad \forall i\colon \neg l_i \Rightarrow x_i' = 0
\end{align*}

The second unimplemented type of constraints are \textbf{reified conditional products} of the form $a(\Pi_{i|l_i=1}x_i)\sim b$. Except for the trivial case with only one factor in the product, these constraints are inherently non-linear, and can not be efficiently converted to a set of linear (in)equalities.

\section{Experiments}
\label{sec:experiments}
In this section we compare \minisatid's performance with CPLEX's performance on problems specified in \fodot. \minisatid takes as input the grounded ECNF constraints, while CPLEX takes as input the MIP translations (in the form of MPS files) returned by the transformation of ECNF explained in the previous section. For six problems, we also verify the performance difference between our automatic translations and a direct MIP model.

\subsection{The benchmark set}
Our benchmark set consists of a mix of Artificial Intelligence and Operations Research problems. Table \ref{tbl:probleminfo} contains an overview.
Problems classified as \emph{search} problems are considered optimization problems with a constant objective function -- each feasible solution is an optimal solution. Search problems are included for two reasons. Firstly, since we are investigating the feasibility of a MIP backend to the \idp system, we are also interested in a MIP solver's performance on non-optimization problems. Secondly, the practical difference between solving a search or optimization problem is often only a relatively simple optimization constraint, compared to other, more crucial problem characteristics.
%\jo{Can give an example: "For example, removing the minimality requirement on the set of colors used for a graph coloring problem does not make it easy to solve when only a couple of colors are given."}

\begin{table}
\vspace{-7 mm}
\caption{Benchmark problem info}
\label{tbl:probleminfo}
    \begin{tabular}{ | p{4.4cm} | c | c | c | c | c |}
    \hline
    Problem & Type & Complexity & \#instances & origin & MIP model \\ \hline
    Assignment & Opt & P & 30 & handmade & - \\
    Shortest Path & Opt & P & 30 (1) & \cite{TSPLIB} & - \\
    Knapsack & Opt & NP-hard & 30 & \cite{kr/LiuJN12} & - \\
    Traveling Salesman (TSP) & Opt & NP-hard & 30 & \cite{TSPLIB} & \cite{MillerTuckerZemlin} \\
    Traveling Umpire (TUP) & Opt & NP-hard & 23 (9) & \cite{TravelingUmpireProblem} & \cite{toffolo2014TUP} \\
    Traveling Tournament (TTP) & Opt & NP-hard & 20 & \cite{TravelingTournament} & - \\
    Maximum Clique & Opt & NP-hard & 30 & \cite{url:fourthasp} & - \\
    Hanoi Four Towers (Hanoi) & Opt & NP-hard & 30 & \cite{url:fourthasp} & handmade \\
    Chromatic Number & Opt & NP-hard & 30 & \cite{TrickResearch} &     \cite{mendez2008cutting} \\
    \hline
    Solitaire & Search & NP-complete & 27 & \cite{url:fourthasp} & - \\
    Permutation Pattern Matching (PPM) & Search & NP-complete & 30 & \cite{url:fourthasp} & - \\
    Graceful Graphs & Search & NP-complete & 30 & \cite{url:fourthasp} & \cite{redl2003graceful} \\
    NQueens Logic & Search & NP-complete & 15 (3) & handmade & handmade \\
    NQueens CP & Search & NP-complete & 15 & handmade & - \\
    \hline
    \end{tabular}
    \vspace{-5 mm}
\end{table}

Note the two different specifications of the NQueens problem: one is logic-based with a boolean variable for each square on the chessboard, and the other is CP-based with a finite-domain variable for each row on the chessboard. %We use both encodings to verify whether \minisatid and CPLEX prefer the same encoding.

\subsection{Evaluating the MIP backend}
We implemented the transformation of ECNF to MIP in version 3.4 of the
\idp system, used the corresponding version of \minisatid as CP
solver, and used IBM ILOG CPLEX Optimization Studio V12.6.0 as MIP
solver, setting MPS as input format. Hardware used is an Intel Core i5-3570 cpu, 8 GiB of ram, running Ubuntu 14.04 64-bit.

ECNF and MPS instances were generated with a 600s 4GB resource limit. Any resource limits reached during the ECNF generation step are classified as unsolved instances, but given between brackets at Table \ref{tbl:probleminfo}. Every ECNF instance generated resulted in a succesfully generated MPS instance.

The solving of these ECNF and MPS instances again had a 600s 4GB resource limit.
Instances are considered succesfully solved if the solver reports the detection of an optimal solution, which for the search category corresponds to any feasible solution.
The benchmark instances and detailed results are publicly available \cite{idp2mip}, while executable specifications can be found at \url{adams.cs.kuleuven.be/idp/idp2mip.html}. Source code for \idp and \minisatid is available upon request.

\begin{figure}
\vspace{-3 mm}
    \centering
    \subfloat[Average ratio of the number of variables and constraints of MIP models compared to the original ECNF theory.]
    {{\includegraphics[height=6.2cm]{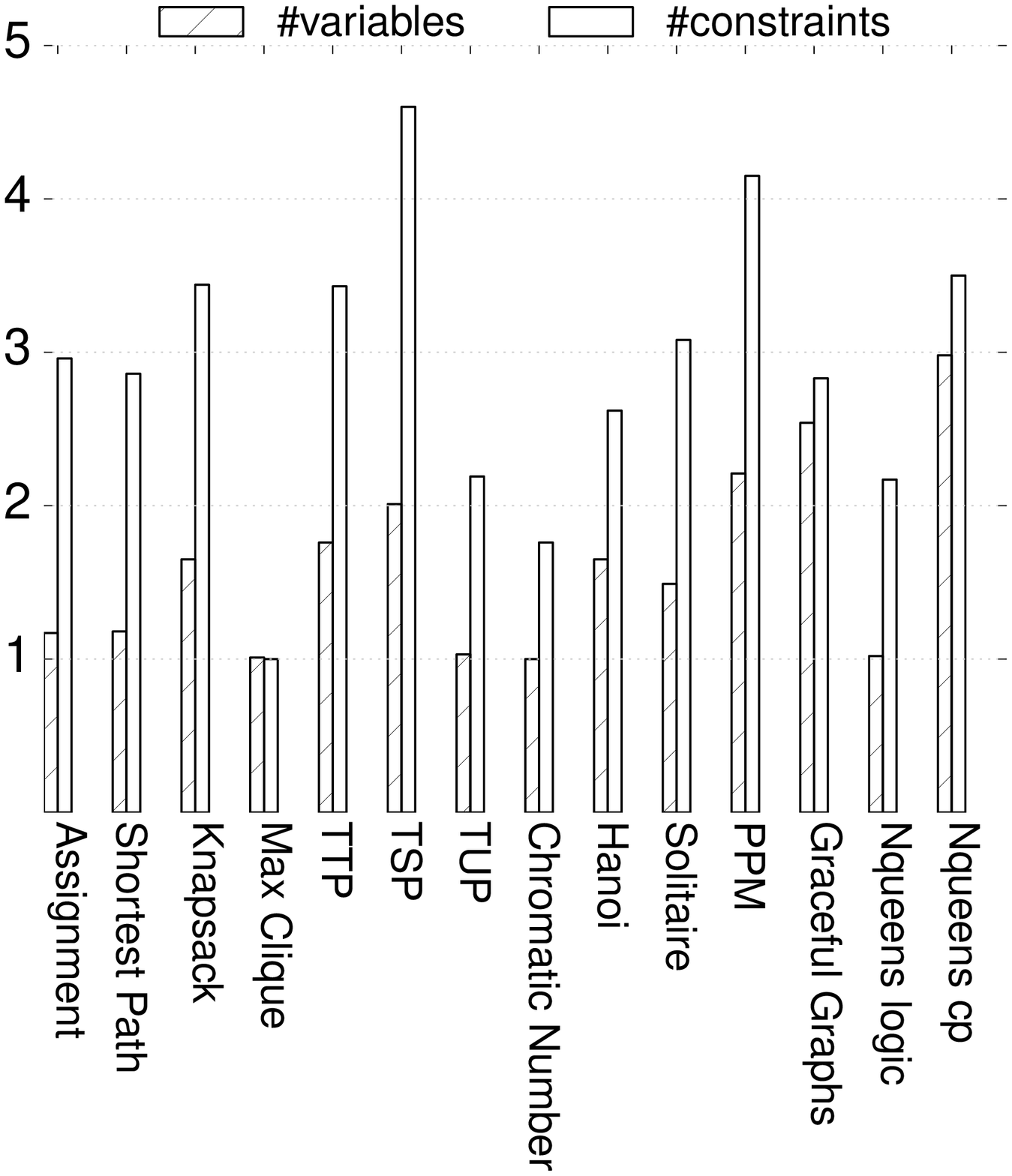} }}%
    \quad
    \subfloat[Percentage of solved instances for \minisatid and CPLEX.]{{\includegraphics[height=6.2cm]{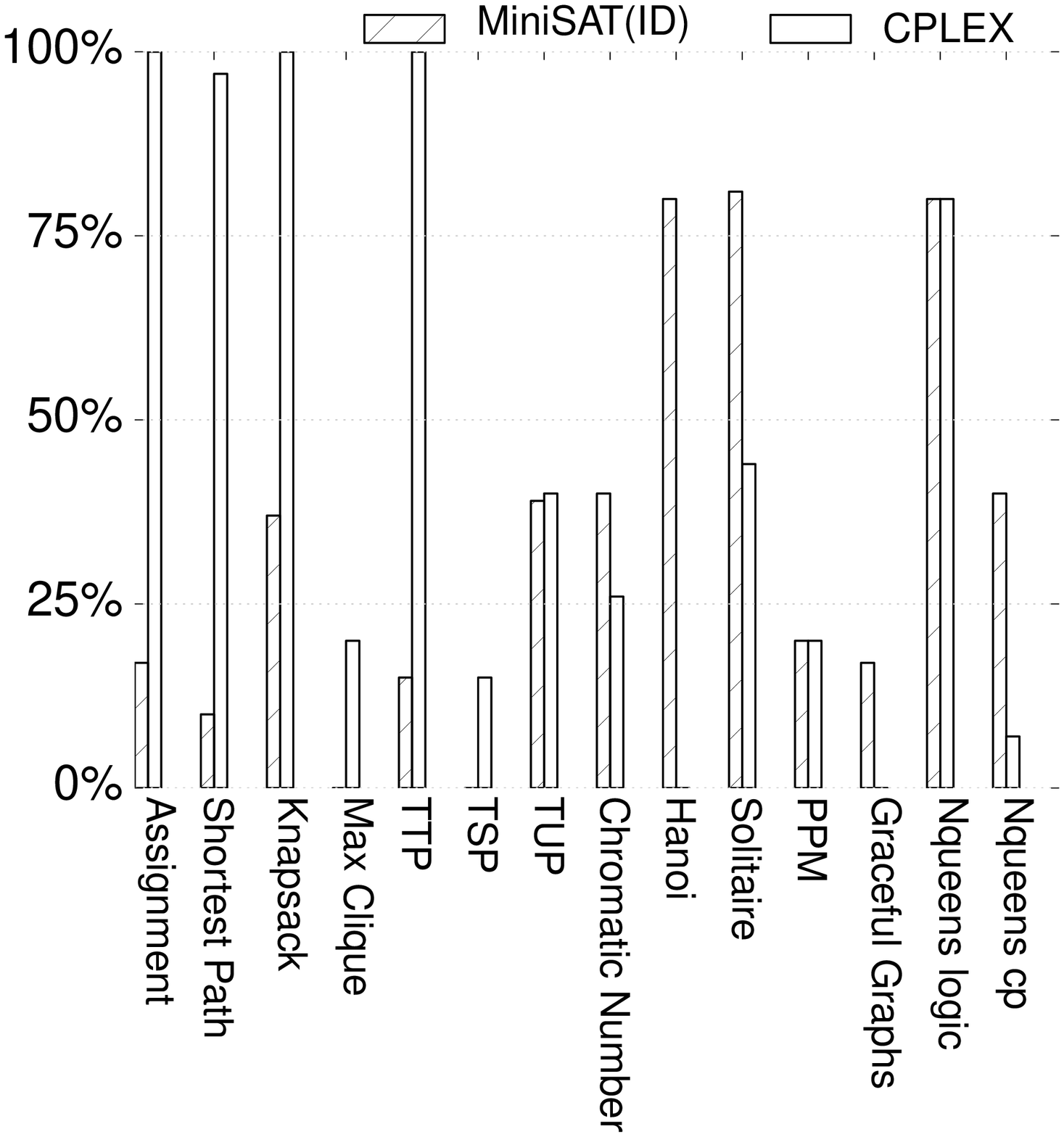}} }
    \vspace{-2 mm}
    \caption{Experimental comparison of \minisatid and MIP backend}
    \label{fig:chart}
    \vspace{-5 mm}
\end{figure}

Figure \ref{fig:chart}(a) shows the average ratio of the size of MIP models compared to the original ECNF theory in terms of the number of variables and constraints. The lower the bars, the less auxiliary variables and constraints are added in the transformed models. Clearly the ratio depends heavily on the problem at hand, or equivalently, on the types of constraints present. For \emph{TSP}, \emph{TTP}, \emph{Solitaire} and \emph{Hanoi}, the level mapping constraints originating from definitions introduce a lot of new variables. For \emph{PPM}, \emph{Graceful Graphs} and \emph{NQueens CP}, the reified linear sums with an $=$ or $\neq$ comparator explain the increased variable count.

Concerning constraint ratios, we counted a definitional rule as a single constraint, since this gives the fairest comparison with the number of constraints (rows) in the transformed MIP model. On average, $2.9$ constraints are present in the transformed model for each constraint in the ECNF theory. This fits the expected variance.

Figure \ref{fig:chart}(b) shows the comparison between the number of optimally solved instances for \minisatid and for CPLEX. For optimization problems, CPLEX almost always outperforms \minisatid. In particular, the large performance difference for the polynomial \emph{Assignment} and \emph{Shortest Path}, and for the weakly NP-hard problem \emph{Knapsack} is impressive. This is explained by the fact that \emph{Assignment} and \emph{Shortest Path} are integer solvable by solving only their linear relaxation, and MIP solvers are known to perform well on \emph{Knapsack}. Our results reflect this, indicating that the transformation from \fodot to MIP did not make these problems harder to solve. For harder problems like \emph{Maximum Clique}, \emph{TTP} or \emph{TSP}, \minisatid can also solve no more than a fraction of CPLEX's solved instances. It is clear that just by switching the backend, the number of solved problems can increase significantly.

As far as \emph{Traveling Umpire} and \emph{Chromatic Number} are concerned, \minisatid performs slightly better than CPLEX, even though these two problems are transformed quite efficiently with few auxiliary variables and constraints. Nonetheless, \emph{Hanoi} is the only clear win for \minisatid, showing that some optimization problems can still be solved better by a strong \emph{explanation} engine than by a strong \emph{relaxation} engine.

Considering the search problems, \minisatid performs better than CPLEX. Firstly, we zoom in on\emph{NQueens logic} and \emph{NQueens CP}. One could have predicted that \emph{NQueens CP} is easier to solve, since the number of variables is apparently smaller than that of \emph{NQueens logic}, and the constraint that there is exactly one queen on a row is satisfied purely by the choice of variables. While we knew that this was not the case for \minisatid, it was surprising to see the same performance discrepancy even more pronounced for CPLEX. The explanation is that \emph{NQueens CP} contains many linear sum with $\neq$ operator constraints, stating that two queens must take a different column and diagonal. As mentioned before, our transformation has bad linear relaxation properties for this type of constraint.
%\san{ya, you're right. We have problem with this remark :D But the remark is totally right for our transformation. The problem is that our transformation is not the best one. 
%Maybe we just have to say something about it in the future work (recent future work actually, I guess that you're also eager to try the "sharp transformation" in our system)
%And I would change the above sentence into: for this constraint, our transformation yeilds bad linear relaxations models.... or something like that :(}
The same holds for the \emph{Graceful Graphs} problem, where all edges must be labeled differently. For this problem, it was known that CP outperforms MIP~\cite{redl2003graceful}.

A third problem where \minisatid performs better is \emph{Solitaire}, a problem similar to \emph{Hanoi} where a goal configuration must be reached starting from an initial configuration using a limited sequence of actions. Since \emph{Solitaire} and \emph{Hanoi} are the only two such problems in our benchmark set, and both are handled unsatisfactory by CPLEX, we hypothesize that explanation based CP solvers, by design, will have an edge over relaxation-focused MIP solvers when solving deep planning problems with large sequences of actions.

%\jo{NOTE NQueens in MIP for other paper...}

\subsection{Comparison with direct MIP models}
\begin{table}[h]
\vspace{-5 mm}
\caption{Solved instances with direct models and transformed models}
\vspace{-3 mm}
\begin{center}
\begin{tabular}{|c|c|c|c|c|c|c|}
	\hline
	& Graceful Graphs & Hanoi & NQueens logic & Chromatic Number & TSP & TUP \\ \hline
Transformed	&		0	&	0	&	12	&6	&	12	&	6 \\ 
Direct	&	0	&	0	&	12	&	15&	13	&	13 \\ \hline
\end{tabular}
\end{center}
\vspace{-9 mm}
\label{tab:direct}
\end{table}
Of course, the results of the previous section can be influenced by the effectiveness of our implemented transformation. To account for this, and to measure the room for improvement in our MIP translation, we compared the transformed MIP model results to reasonably optimized direct MIP models. These direct MIP models were either taken from literature, or constructed by hand, with the exact origin presented in the last column of Table \ref{tbl:probleminfo}. To keep the comparison to a ground-and-solve approach fair, we only took MIP models that were ``simple'' in the sense that no column generation, decomposition approaches or special cut generation algorithms were used. In essence, the model must be solvable with a classic branch-and-bound algorithm.

The results are available in Table \ref{tab:direct}. Firstly, it is clear that both \emph{Hanoi} and \emph{Graceful Graphs} show no improvement, showing that these problems remain hard, even with a more optimized MIP model. Secondly, our transformation still leaves room for improvement. For example, the direct \emph{Chromatic Number} model utilizes the direction of the optimization constraint to drop superfluous constraints enforced by the direction of optimization.

\section{Conclusion and future work}
\label{sec:conclusion}
This paper confirms that the \idp system is no exception to other systems -- 
a MIP backend allows \idp to solve a great range of extra problems.
However, MIP cannot completely replace the CP backend. Certain deep 
planning problems are not satisfactory solved, and the lack of tightness for 
translations of reified linear sums with $\neq$ or $=$ operators poses problems.

We also point out that integrality constraints on level mapping variables should
be dropped. Lastly, we do not agree with the sentiment that for 
problems involving only binary variables and constraints, the 
performance of CPLEX is not as competitive \cite{kr/LiuJN12}. Our experiments
show CPLEX' performance on boolean problems such as \emph{Maximum Clique}, 
\emph{NQueens logic} and \emph{Chromatic Numbering} to be more than competitive.

As far as future work is concerned, it would be interesting to improve the tightness 
of the transformation for reified linear sums with $\neq$ or $=$ operators. Inspiration can be drawn from \cite{2000Refalo}. Also, we plan to 
implement a linear relaxation propagator in \minisatid, possibly borrowing ideas 
from the SAT Modulo Theories community~\cite{lncs/dutertredemoura}.

\section*{Acknowledgement}
Work supported by the Belgian Science Policy Office (BELSPO) in the Interuniversity Attraction Pole COMEX. (http://comex.ulb.ac.be), 
Research Foundation Flanders (FWO), BOF (GOA De Raedt),
and the Marie  Curie ITN STEEP (Grant Agreement no. 316560, http://www.steep-itn.eu/steep/index.aspx).

\bibliography{LION,idp-latex/krrlib}
\bibliographystyle{splncs03}

\end{document}